\documentclass[journal]{IEEEtran}
\ifCLASSINFOpdf
\else
\fi
\usepackage{graphicx}
\usepackage{bm}
\usepackage{algorithm}
\usepackage{algorithmic}
\usepackage{color}
\usepackage{mathrsfs}
\usepackage{multirow}
\usepackage{makecell}
\usepackage{amsmath}

\begin{document}
%
\title{Progressive Bilateral-Context Driven Model for Post-Processing Person Re-Identification}
%
%
%

\author{Min~Cao, Chen~Chen, Hao~Dou, Xiyuan~Hu, Silong~Peng
        and~Arjan~Kuijper

\thanks{Chen Chen is the corresponding author (e-mail: chen.chen@ia.ac.cn).}
\thanks{Min Cao is with School of Computer Science and Technology, Soochow University, 215006 Suzhou, China. She is also with the Institute of Automation Chinese Academy of Sciences Beijing 100190, China and Fraunhofer IGD, 64283 Darmstadt, Germany (e-mail: mcao@suda.edu.cn).}
\thanks{Chen Chen, Hao Dou and Silong Peng are with the Institute of Automation Chinese Academy of Sciences Beijing 100190, China.}
\thanks{Xiyuan Hu is with School of Computer Science and Engineering, Nanjing University of Science and Technology.}
\thanks{Arjan Kuijper is with the Mathematical and Applied Visual Computing, TU Darmstadt, 64283 Darmstadt, Germany, and Fraunhofer IGD, 64283 Darmstadt, Germany.}}

\maketitle

\begin{abstract}
Most existing person re-identification methods compute pairwise similarity by extracting robust visual features and learning the discriminative metric. Owing to visual ambiguities, these content-based methods that determine the pairwise relationship only based on the similarity between them, inevitably produce a suboptimal ranking list. Instead, the pairwise similarity can be estimated more accurately along the geodesic path of the underlying data manifold by exploring the rich contextual information of the sample. 
In this paper, we propose a lightweight post-processing person re-identification method in which the pairwise measure is determined by the relationship between the sample and the counterpart's context in an unsupervised way. We translate the point-to-point comparison into the bilateral point-to-set comparison. The sample's context is composed of its neighbor samples with 
two different definition ways: the first order context and the second order context, which are used to compute the pairwise similarity in sequence, resulting in a progressive post-processing model. 
The experiments on four large-scale person re-identification benchmark datasets indicate that (1) the proposed method can consistently achieve higher accuracies by serving as a post-processing procedure after the content-based person re-identification methods, showing its state-of-the-art results, (2) the proposed lightweight method only needs about 6 milliseconds for optimizing the ranking results of one sample, showing its high-efficiency. 
Code is available at: https://github.com/123ci/PBCmodel.
\end{abstract}

\begin{IEEEkeywords}
Person re-identification, post-processing, contextual information.
\end{IEEEkeywords}

%
\IEEEpeerreviewmaketitle

\section{Introduction}
\label{i}

Person re-identification (re-id) \cite{karanam2019a} aims to identify designated individuals from a large amount of pedestrian images across non-overlapping camera views and is a critical task for the realization of an intelligent video monitoring system. Due to changes of the visual appearance of a person caused by different illumination, poses and background across camera views, person re-id is a very challenging task.

\begin{figure}
\begin{center}
   \includegraphics[width=1\linewidth, height=0.33\linewidth]{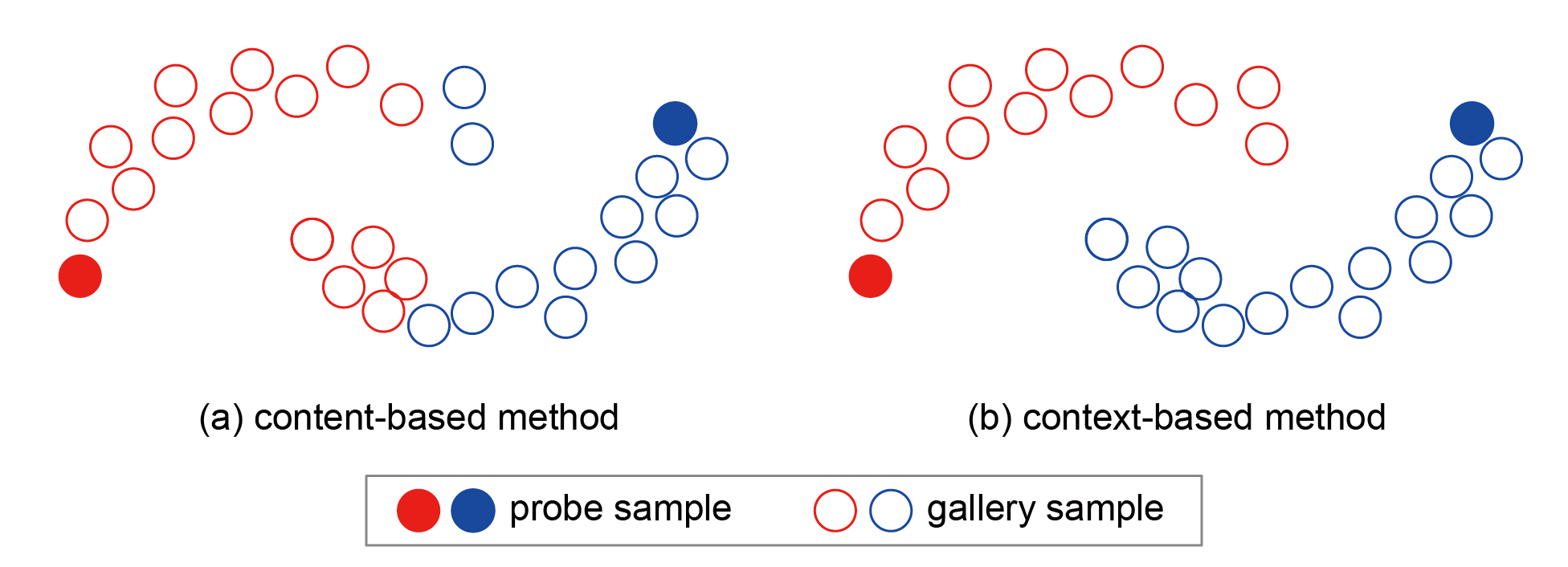}
\end{center}
   \caption{Comparison of retrieval results produced by (a) content-based method and (b) context-based method. The color of each gallery sample is marked according to the similarity to the probe sample. Better viewed in color.}
\label{fig1}
\end{figure}

Essentially, person re-id can be regarded as a retrieval task. Given a probe sample and a collection of gallery samples, our aim is to compute the probe sample's ranking list, in which the positive gallery samples with the same identity as the probe sample are expected to be closer to the top of the ranking list. Currently, a mainstream solution \cite{matsukawa2016hierarchical,zhao2018person,saquib2018pose,lin2017learning} is to learn a discriminative feature mapping based on the extracted sample image features in the training set, and compute the similarity of the newly learned feature vectors between the probe sample and each of gallery samples in Euclidean space. The ranking list is obtained based on the pairwise similarities in descending order. It is a classic content-based image retrieval (CBIR) system in which the relationship between the probe sample and gallery sample is determined only by the similarity between them. This has been demonstrated \cite{donoser2013diffusion} to be insufficient to reveal the structure of the dataset manifold. 
Instead, the pairwise similarity can be computed more accurately by taking into account the contextual information of sample and the structure of the dataset manifold. 
A toy example in Fig.~\ref{fig1} illustrates the concept. Compared to the content-based method with the Euclidean distance for retrieval, the context-based method considers the geometry of data manifold and produces correct retrieval result.

Several context-based person re-id methods have been proposed recently \cite{yan2019learning,luo2019spectral,bai2017scalable,garcia2015person,saquib2018pose,wang2019incremental}, in which the contextual information of samples is introduced to assist the computation of the pairwise similarity by the ranking list comparison \cite{zhong2017re,leng2015person,bai2016sparse,wang2016zero-shot}, or by building an end-to-end deep learning framework based on graph theory \cite{shen2018person,shen2018deep}. 
In these methods, the samples in one batch or dataset are utilized as the context sample to assist person re-id, and the set-to-set comparison or graph theory based deep learning are made for person re-id. The useful context samples are not fully explored due to the indiscriminate use of samples, and the computation complexity is comparatively high due to the computation of the set-to-set comparison and the need for training model.

\begin{figure}
\begin{center}
   \includegraphics[width=1\linewidth, height=0.37\linewidth]{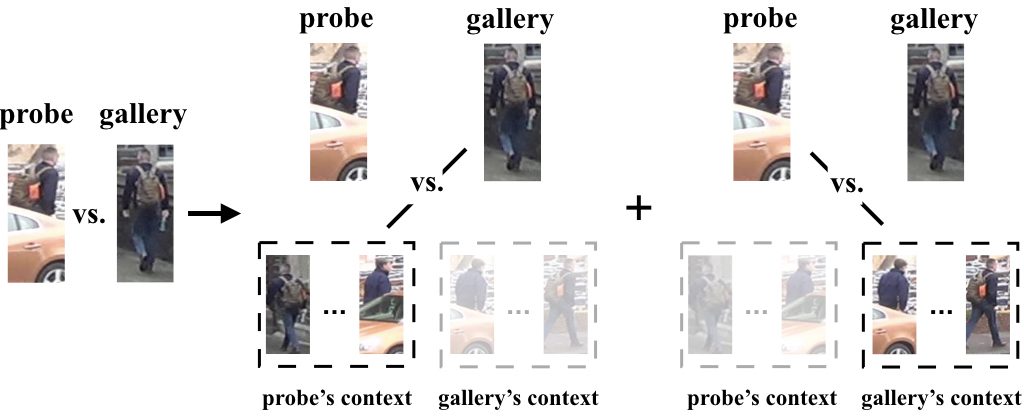}
\end{center}
   \caption{Illustration of the gist of the proposed method.}
\label{fig8}
\end{figure}

In this paper, we propose a novel context-driven post-processing person re-id method. The gist of the proposed method is shown in Fig.~\ref{fig8}, the similarity between two target samples is computed by measuring the relationship between the target sample and the counterpart's context. 
The point-to-point comparison is translated into the bilateral point-to-set comparison with low computation complexity. 
The neighbor samples of the target samples provide the reliable and effective contextual information, so we define the target sample's context by the k-nearest neighbors (k-nn) algorithm. We propose two ways of computing the context: the first order k-nn and the second order k-nn, and obtain the first order context and the second order context, respectively. 
Compared to the first order context, the second order context considering high-order information provides more reliable and effective contextual information. Both of contexts are adopted to assist the computation of the pairwise similarity in sequence, resulting in a progressive process of computation. The contextual information of sample is adequately explored for assisting person re-id.

As a post-processing method, the method with a low time and space complexity is very necessary to apply to the large scale real-world scenario. 
Most of post-processing methods \cite{zhong2017re,saquib2018pose,luo2019spectral,leng2015person} improve the performance by the ranking list comparison, i.e.\ a set-to-set comparison, or by optimizing the ranking results of all gallery samples, or by utilizing both probe and gallery samples in datasets. Among them, some methods need to train a model in the supervised setting. All these result in the high time and space complexity of the method. 
By contrast, in the proposed method, 
(1) the post-processing is made with a bilateral point-to-set comparison mechanism in an unsupervised setting;
(2) the post-processing is made for a fraction of gallery samples at the top of each probe sample's initial ranking list, since only the top positions of the ranking list are likely to contain the most relevant samples as to the probe sample; 
(3) the post-processing is made only by utilizing the gallery samples in the dataset to assist person re-id, meeting the reality in which the input of person re-id system is only a probe sample and a collection of gallery samples.
In addition, since only the gallery samples are utilized in the proposed method, the computation related to the gallery samples can be done offline and there are only some simple retrieve operations related to the probe sample in online process. 
As a result, the time and space complexity of the proposed method is very low. 

In summary, our contributions can be summarized as follows: 
(1) We propose to compute the pairwise similarity by a bilateral point-to-set comparison. Compared with the mainstream context-based methods with the set-to-set comparison, the proposed bilateral-context driven model with the bilateral point-to-set comparison achieves the state-of-the-art accuracies under lower computation complexity.
(2) We propose a progressive post-processing strategy by adopting the first order context and second order context in sequence to optimize the ranking results. The contextual information is fully utilized for assisting person re-identification.
(3) The application in large scale real-world scenario is well considered in the proposed method. We develop the effective designs including the gallery based offline computation and the top ranked sample optimization, realizing a lightweight and practical post-processing re-identification method.

We conduct extensive experiments on four large-scale person re-id datasets, demonstrating the effectiveness, high efficiency and robustness of the proposed method.

\section{Related work}
Because of the promising application prospects, person re-id has attracted more and more attention in the academic community in recent years. For a detailed review, the interested readers are referred to \cite{karanam2019a}. In this paper, we propose to improve the performance of person re-id by making full use of the contextual information of sample. The proposed method can serve as a post-processing method to refine the results from other person re-id methods. In this section we thus introduce related work on post-processing person re-id methods and context-based person re-id methods, respectively.

\subsection{Post-processing person re-id methods} 
To increase the performance of person re-id,  post-processing methods have recently gained attention. 
These methods are applied after the content-based methods, and boost the performance of person re-id by (1) the human feedback on-the-fly to interactively update re-id model \cite{liu2013pop,wang2014region,wang2016human,liu2019deep} or (2) automatically exploring rich contextual information encoded in the relations among several images \cite{ye2016person,bai2019re-ranking,zhong2017re,saquib2018pose,bai2017scalable,leng2015person,wang2016zero-shot}.
Compared to the feedback based post-processing methods, the context-based methods have wider applicability due to without any manual intervention. 
Ye et al.\ \cite{ye2016person} proposed to optimize the initial ranking list by pulling the similarity relationship and pushing the dissimilarity relationship based on multi content-based methods. 
Bai et al.\ \cite{bai2019re-ranking} concentrated on the ensemble of multiple metrics for the post-processing of person re-id. They are, however, inconvenient to apply in real-world person re-id, since the initial ranking lists of the multi content-based methods are required in these methods.
Zhong et al.\ \cite{zhong2017re} refined the pairwise similarity in virtue of the similarities of k-reciprocal nearest neighbors between the probe sample and the gallery sample. Similarly, in \cite{saquib2018pose}, Expanded Cross Neighborhood (ECN) distance with the ranking list comparison was proposed, and in \cite{leng2015person}, the initial ranking list was refined based on the assumption that positive sample pairs possess similar k-nearest neighbors, refer to context similarity. The above three methods are related to the ranking list comparison with high computation complexity.
And in the process of calculation in \cite{zhong2017re,saquib2018pose}, all samples in both probe set and gallery set are used to assist the optimization of ranking list, which is not conform with reality where only one probe image and a collection of gallery samples are given in most cases. 
Bai et al.\ \cite{bai2017scalable} proposed to boost the performance by manifold-based affinity learning on the training set, which has a negative impact on the time and space complexity of algorithm, because of the need for training a model. 
In \cite{Zheng2018Hetero}, the pairwise similarity is measured by three order random walks on the hetero-manifold: from the probe to its neighbor, from the probe's neighbor to the gallery's neighbor, and from the gallery's neighbor to the gallery. Essentially, it is similar to the methods \cite{zhong2017re,saquib2018pose,leng2015person} in which the pairwise similarity is computed by the set-to-set comparison.

Compared with the post-processing methods mentioned above, the proposed method (1) achieves the performance enhancement with no human feedback and based on the initial ranking results from one content-based method, (2) doesn't involve the comparison between the ranking lists, (3) rests on the assumption that one probe sample and a collection of gallery samples are only given in the person re-id system considering the practicality, and (4) is solved in an unsupervised way. Therefore, the proposed method is more in line with the real-world large scale person re-id. 

\subsection{Context-based person re-id methods} 
Besides the context-based post-processing person re-id methods mentioned above, some context-based learning methods are proposed in recent years, in which the contextual information is utilized based on graph theory with an end-to-end deep learning framework \cite{yan2019learning,shen2018person,shen2018deep,luo2019spectral}. By contrast, the proposed method has a good compatibility and high efficiency, and can utilize various content-based methods as baseline and enhance the performance of person re-id by the contextual information in the unsupervised setting with low computation complexity. It can be conveniently applied in large scale real-world scenario. 

\section{Proposed method}
Given a probe sample $p$ and a collection of gallery samples $G=\left \{ g_i\:| \: i=1,\cdots ,N\right \}$, where $N$ denotes the number of gallery samples, an initial ranking list $\mathcal{R}^o_p=\left \{ g^o_{1},g^o_{2}, \cdots ,g^o_{N}\right \}$ can be obtained by a content-based person re-id method as the baseline, with $S^o_{p,g^o_1}> S^o_{p,g^o_2}\cdots >S^o_{p,g^o_N}$, where $g^o_i\in G$ ($i=1,2,\cdots, N$) and $S^o_{p,g^o_i}$ denotes the similarity between $p$ and $g^o_i$.
The initial ranking list $\mathcal{R}^o_p$ is a suboptimal solution, since the image pairs are matched only based on the similarity relation between these two individuals and the rich information on the similarity relation with the other samples is not fully explored. 

In view of this, we optimize the initial results $\mathcal{R}^o_p$ by re-compute the pairwise similarities between $p$ and each gallery sample $g_{i}$ ($i=1,2,\cdots, N$) with the help of the bilateral-contextual information, i.e.\ the contexts of both $p$ and $g_{i}$, so that more positive gallery samples rank at the top of the list and improve the accuracy of the person re-id. 
We obtain two types of contexts based on the first order k-nn and the second order k-nn, respectively. The second order context is firstly used for optimizing the initial results, then on this basis, the first order context is used to further fine-tune the results, realizing a progressive optimization. 

Without loss of generality, in the following, two target samples are denoted as the probe sample $p$ and the gallery sample $g$. We firstly introduce the definition of target sample's context, then we compute the similarity between the target sample and the counterpart’s context. Finally, we introduce the proposed progressive bilateral-context driven optimization for the initial results $\mathcal{R}^o_p$. 

\subsection{The definition of context}
\label{cc}
The target sample's context is composed of the neighbor samples. Thus we define the context with the aid of the k-nn algorithm. 

In practice, the target sample's context is easily obtained based on the initial ranking list, i.e.\ the top-$k$ samples of the ranking list as the context. Formally, for the probe sample $p$, we define its first order context as the sample set $\mathcal{C}^1(p,k)$ such that
\begin{equation}\label{e1}
\mathcal{C}^1(p,k)=\left \{ g^o_{1},g^o_{2}, \cdots ,g^o_{k}\right \}, \\
\end{equation}
where 
$\left | \mathcal{C}^1(p,k) \right |=k$, and $\left | \cdot  \right |$ denotes the number of samples in the set. Similarly, the context $\mathcal{C}^1(g,k)$ of gallery $g$ can be obtained based on the ranking list $\mathcal{R}_g$. Note that $\mathcal{R}_g$ is the permutation and combination among the gallery samples $g_i$ ($i=1,\cdots ,N$) and is computed based on the baseline method.

Furthermore, for obtaining more contextual information, naturally we need to set a larger value of $k$. However, more and more interference samples will be introduced with increase of $k$. To exploit more reliable and effective contextual information, we consider high-order based neighbor samples and propose a second order context. Specifically, the second order context of the probe sample $p$ is defined as
\begin{equation}\label{e2}
\mathcal{C}^2(p,k_0,k)=\left \{\mathcal{C}^1(g^o_1,k),\mathcal{C}^1(g^o_2,k),\cdots , \mathcal{C}^1(g^o_{k_0},k)\right \}, 
\end{equation}
where $\left | \mathcal{C}^2(p,k_0,k) \right |=k_0\times k$. The second order context is made up of the first order contexts of top-$k_0$ samples in the ranking list of $p$. A similar context set $\mathcal{C}^2(g,k_0,k)$ can be obtained for the gallery $g$. 

\begin{figure}
\begin{center}
   \includegraphics[width=0.9\linewidth, height=0.65\linewidth]{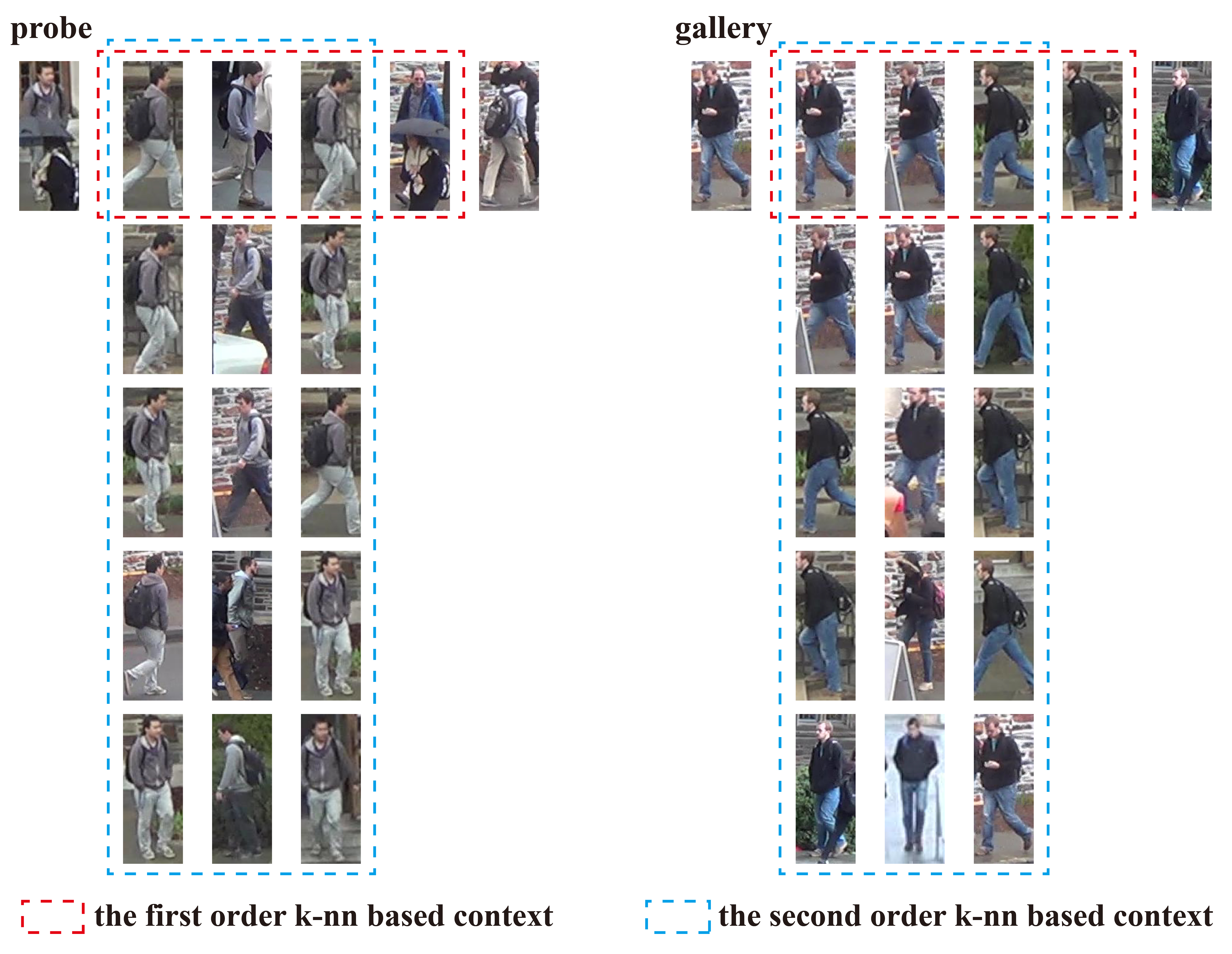}
\end{center}
   \caption{Illustration of the first order context and the second order context with $k_0=2$ and $k=3$ on the person re-id DukeMTMC dataset \cite{ristani2016performance}. Better viewed in color.}
\label{fig3}
\end{figure}

For a better understanding, Fig.~\ref{fig3} illustrates the computation of the first order context and second order context, respectively. It can be seen that there are samples with similar appearance to the target sample in the context. In processing the computation of the pairwise similarity, the introduction of these samples implicitly expands the amount of effective  information and will have a positive effect on improving the accuracy of the person re-id. It is noteworthy that the amount of effective information in the second k-nn context mainly depends on the top-$k_0$ samples. If one of the top-$k_0$ samples is not similar to the target sample, there are $k$ interference samples introduced into its second order context. To ensure the quality of the context, a small $k_0$ should be given, in contrast, a little larger value can be set for $k$. 

For readability purposes, $\mathcal{C}_p=\left \{\rho_1, \rho_2, \cdots, \rho_{\left | \mathcal{C}_{p} \right |}\right \}$ and $\mathcal{C}_{g}=\left \{\varrho_1, \varrho_2, \cdots, \varrho_{\left | \mathcal{C}_{g} \right |}\right \}$ are used to generally refer to the context of $p$ and the context of $g$ in the following, respectively.

\subsection{The similarity between target sample and counterpart's context}
\label{ssc}

For the target samples $p$ and $g$, the corresponding contexts $\mathcal{C}_p$ and $\mathcal{C}_{g}$ are computed according to Section~\ref{cc}. In this subsection, we focus on the computation of the similarity between the target sample and the counterpart's context.

Specifically, the similarity between the target sample and the counterpart's context is defined as:
\begin{equation}\label{e14}
S_{p,\mathcal{C}_g}= \sum_{i=1}^{\left | \mathcal{C}_g \right |}w_{\varrho_i}\cdot r_{p,\varrho_i},
\end{equation}
\begin{equation}\label{e13}
S_{g,\mathcal{C}_p}= \sum_{i=1}^{\left | \mathcal{C}_p \right |}w_{\rho_i}\cdot r_{g,\rho_i},
\end{equation}
where $r_{p,\varrho_i}$ and $r_{g,\rho_i}$ denote the similarity between the target sample and sample in counterpart's context, $w_{\varrho_i}$ and $w_{\rho_i}$ denote the weight of the context sample.

Next, we provide detailed description for the computations of similarity value $r_{g,\rho_i}$ and weight $w_{\rho_i}$. The computations of $r_{p,\varrho_i}$ and $w_{\varrho_i}$ can be done similarly and are not described in the following. 

\subsubsection{The computation of similarity value $r_{g,\rho_i}$}
\label{csv}
Naturally, this similarity relationship should be described based on the matching degree of the appearance between the samples. A straightforward way is to use the matching value. However, this would lead to that the method's performance has high sensitivity to the variation of the matching value. Actually, the ranking positions from each other’s ranking list are more reliable and more effective information to describe the matching degree between the target sample and the sample in the counterpart‘s context. Therefore, we use the rank-based criterion to measure the similarity.

Given the ranking list $\mathcal{R}_g$ of the gallery sample $g$, we locate the ranking position $l _{\rho_i}(\mathcal{R}_g)$ of the sample $\rho_i\in \mathcal{C}_{p}$ ($i=1,2,\cdots, \left | \mathcal{C}_{p} \right |$) in the ranking list $\mathcal{R}_g$. The similarity between the gallery sample $g$ and the sample $\rho_i$ in the context of probe sample $p$ is defined as
\begin{equation}\label{e3}
r_{g,\rho_i}=\frac{1}{l _{\rho_i}(\mathcal{R}_g)}.
\end{equation}

However, $\rho_i$ ranking well in the ranking list $\mathcal{R}_g$ does not always represent that $g$ ranks well in the ranking list $\mathcal{R}_{\rho_i}$. It is not reliable to measure the similarity only based on the nonreciprocal rank information by Eq.\ (\ref{e3}). We deem that $g$ and $\rho_i$ are similar to each other if they all rank well in the ranking lists of each other. For that reason, the similarity between $g$ and $\rho_i$ can be computed by the reciprocal rank-based measure such that
\begin{equation}\label{e4}
r_{g,\rho_i}=\frac{1}{max(l _{\rho_i}(\mathcal{R}_g),l _{g}(\mathcal{R}_{\rho_i}))},
\end{equation}
or
\begin{equation}\label{e5}
r_{g,\rho_i}=\frac{1}{l _{\rho_i}(\mathcal{R}_g)+l _{g}(\mathcal{R}_{\rho_i})},
\end{equation}
Compared to the similarity measure in Eq.\ (\ref{e3}), the measures in Eq.\ (\ref{e4}) and Eq.\ (\ref{e5}) consider the reciprocal rank information and satisfy the symmetry, i.e.\ $r_{g,\rho_i}=r_{\rho_i,g}$.

However, there are some cases where the result will be a biased one if we only adopt the reciprocal rank-based measure in Eq.\ (\ref{e4}) or in Eq.\ (\ref{e5}) to compute the similarity. For example, for the sample $\rho_1$ we obtain $l _{\rho_1}(\mathcal{R}_g)=1$ and $l _{g}(\mathcal{R}_{\rho_1})=3$, and for the sample $\rho_2$ we obtain $l _{\rho_2}(\mathcal{R}_g)=2$ and $l _{g}(\mathcal{R}_{\rho_2})=3$. It is clear that $g$ is more similar to the sample $\rho_1$ than to the sample $\rho_2$ according the ranking positions. However if we compute the similarity according to Eq.\ (\ref{e4}), $r_{g,\rho_1}=r_{g,\rho_2}=\frac{1}{3}$ and these two samples $\rho_1$ and $\rho_2$ have the same similarity values to the gallery sample $g$. By contrast, the similarity measure is accurate based on Eq.\ (\ref{e5}). For another example, for the sample $\rho_1$ we obtain $l _{\rho_1}(\mathcal{R}_g)=1$ and $l _{g}(\mathcal{R}_{\rho_1})=7$, and for the sample $\rho_2$ we obtain $l _{\rho_2}(\mathcal{R}_g)=4$ and $l _{g}(\mathcal{R}_{\rho_2})=4$. We compute $r_{g,\rho_1}=r_{g,\rho_2}=\frac{1}{8}$ according to Eq.\ (\ref{e5}), which shows the same similarity of $\rho_1$ and $\rho_2$ to $g$. In fact, compared with the pair $g$ and $\rho_2$, there is a big difference of ranking positions for the pair $g$ and $\rho_1$, which indicates the instability of similarity relation and a smaller similarity value should be given. This drawback can effectively be avoided by computing the similarity based on Eq.\ (\ref{e4}).

According to the above discussion, we thus compute the similarity between the gallery sample $g$ and the sample $\rho_i$ in the context of the probe sample $p$ by combining the reciprocal rank-based similarity measures in Eq.\ (\ref{e4}) and in Eq.\ (\ref{e5}):
\begin{equation}\label{e6}
r_{g,\rho_i}=\frac{1}{l _{\rho_i}(\mathcal{R}_g)+l _{g}(\mathcal{R}_{\rho_i})+max(l _{\rho_i}(\mathcal{R}_g),l _{g}(\mathcal{R}_{\rho_i}))}.
\end{equation}

\subsubsection{The computation of weight $w_{\rho_i}$}
\label{wsc}
 
The context of the target sample is exploited as intermediary to assist the computation of the pairwise similarity. For the samples in the target sample's context that are more similar to the target sample, they are regarded as more reliable intermediary samples, since they share visual appearance with more similarity to the target sample and provide more valuable information, so they should be assigned to the larger weights during the computation of the pairwise similarity. In view of this, we assign different weights to each of samples in the context according to its similarity to the target sample. 

Next, we introduce the computation of sample's weight in the first order context and the second order context, respectively.

For the samples $\rho_i$ ($i=1,2,\cdots ,k$) in the first order context $\mathcal{C}^1(p,k)$ of the probe sample $p$, we compute the weight based on the similarity between $\rho_i$ and $p$, measured by the ranking positions of each other's ranking lists,
\begin{equation}\label{e7}
\begin{aligned}
w_{\rho_i}=\frac{1}{l _{\rho_i}(\mathcal{R}_p)+l _{p}(\mathcal{R}_{\rho_i})+max(l _{\rho_i}(\mathcal{R}_p),l _{p}(\mathcal{R}_{\rho_i}))}, \\
\rho_i\in \mathcal{C}^1(p,k)
\end{aligned}
\end{equation} 

For the samples $\rho_i$ ($i=1,2,\cdots ,k_0\times k$) in the second order context $\mathcal{C}^2(p,k_0,k)$ of the probe sample $p$, different weights have been assigned based on the definition in Eq.\ (\ref{e2}). If a sample $\rho_i$ is similar to $p$, it is likely that $\rho_i$ not only belongs to $\mathcal{C}^1(g^o_m,k)$, but also to $\mathcal{C}^1(g^o_n,k)$ ($m\neq n < k_0$). There will be multiple occurrences of this person in $\mathcal{C}^2(p,k_0,k)$ according to Eq.\ (\ref{e2}) that can also easily be observed in Fig.~\ref{fig3}. As a result, compared to other samples with the number of occurrences less than the sample $\rho_i$, the information of $\rho_i$ is utilized multiple times when $\mathcal{C}^2(p,k_0,k)$ is used to assist the re-id, and the larger weight is implicitly given to the sample $\rho_i$. To sum up, in the second order context, the number of the sample's occurrences in context reflects the degree of similarity between this sample and the target sample, and is also a direct expression of the weight.

Furthermore, for obtaining more accurate weight values of samples in the second order context, we introduce the reliability measure of the ranking list into the second order context. The second order context is made up of the $k_0$ first order contexts $\mathcal{C}^1(g^o_j,k)$ ($j=1,2,\cdots,k_0$) obtained by the ranking lists of the sample $g^o_j$. By measuring the reliability of these ranking lists, we can further measure the similarity between $p$ and $\rho_i$ and thereby obtain the weight of $\rho_i$. Specifically, the reliability of the sample $g^o_j$'s ranking list reflects to some extent the degree of similarity between the sample $g^o_j$ and the sample $\rho_i \in \mathcal{C}^1(g^o_j,k)$ in the second order context. The more reliable the sample $g^o_j$'s ranking list is, the more similar the $g^o_j$ and samples in $\mathcal{C}^1(g^o_j,k)$ are.
For Section~\ref{cc} we know that a small $k_0$ is given for ensuring high similarities between $p$ and these $k_0$ samples $g^o_j$ ($j=1,2,\cdots,k_0$). Therefore, the ranking list of the sample $g^o_j$ is more reliable, the samples belonging to set $\mathcal{C}^1(g^o_j,k)$ in the second order context are more similar to $p$ and should be assigned to the larger weights. 

Inspired by \cite{pedronette2012exploiting}, we measure the reliability of a ranking list based on the cohesion of top-$k$ samples in this ranking list. For readability purposes, assuming a sample $q$ is given and we compute the ranking list's reliability of $q$. Let $\widetilde{\mathcal{R}}_{q}=\left \{ q_{1},q_{2}, \cdots ,q_{k}\right \}$ be a set with top-$k$ samples of the ranking list $\mathcal{R}_{q}$, and $\widetilde{\mathcal{R}}_{q_i}=\left \{ q_{i1},q_{i2}, \cdots ,q_{ik}\right \}$ be a set with top-$k$ samples of the ranking list $\mathcal{R}_{q_i}$ ($q_i\in \widetilde{\mathcal{R}}_{q}$, $i=1,2,\cdots,k$). The reliability of the sample $q$'s ranking list is measured as follows:
\begin{equation}\label{e8}
\kappa_{q}=\frac{\sum _{q_i\in \widetilde{\mathcal{R}}_{q}}\sum _{q_{ij}\in \widetilde{\mathcal{R}}_{q_i}}l _{q_i}({\mathcal{R}}_{q})\mathbf{1}_{q_{ij}}(\widetilde{\mathcal{R}}_{q})}{\sum _{q_i\in \widetilde{\mathcal{R}}_{q}}\sum _{q_{ij}\in \widetilde{\mathcal{R}}_{q_i}}l _{q_i}({\mathcal{R}}_q)},
\end{equation}
where $l _{q_i}({\mathcal{R}}_q)=\frac{1}{i}$ is the ranking position of sample $q_i$ in ranking list ${\mathcal{R}}_{q}$. $\mathbf{1}_{q_{ij}}(\widetilde{\mathcal{R}}_{q})$ is an indicator function and determines if the sample $q_{ij}$ belonging to $\widetilde{\mathcal{R}}_{q_i}$ also belongs to $\widetilde{\mathcal{R}}_{q}$:
\begin{equation}\label{e9}
\mathbf{1}_{q_{ij}}(\widetilde{\mathcal{R}}_{q}) = \left\{ {\begin{array}{*{20}{c}}
   1 & q_{ij}\in \widetilde{\mathcal{R}}_{q}  \\
   0 & q_{ij}\notin \widetilde{\mathcal{R}}_{q}  \\
\end{array}} \right.
\end{equation}

The value of the reliability ranges from $0$ to $1$. If all samples $q_{ij}$ in $\widetilde{\mathcal{R}}_{q_i}$ ($i=1,2,\cdots,k$) belong to the set $\widetilde{\mathcal{R}}_{q}$, $\kappa_{q}=1$. It indicates a perfect cohesion of top-$k$ samples in the ranking list $\mathcal{R}_{q}$, which shows the excellent reliability of the sample $q$'s ranking list. 

According to Eq.\ (\ref{e8}), we compute the reliability of the ranking list of the samples $g^o_j$ ($j=1,2,\cdots,k_0$), denoted as $\kappa_{g^o_j}$ ($j=1,2,\cdots,k_0$). Then, the weight of the sample $\rho_i$ in the second order context $\mathcal{C}^2(p,k_0,k)$ of the probe sample $p$ is given as
\begin{equation}\label{e10}
w_{\rho_i}=\kappa_{g^o_j}, \;\; \rho_i\in \mathcal{C}^1(g^o_j,k) \subseteq \mathcal{C}^2(p, k_0, k), \; j=1,2,\cdots,k_0
\end{equation}

\subsection{Progressive bilateral-context driven optimization}
We optimize the initial ranking list from the content-based method by re-computing the pairwise similarity with the aid of the context. 

\subsubsection{Bilateral-context}
We compute the similarity between $p$ and $g$ by considering both the relation between the probe sample $p$ and the gallery sample's context $\mathcal{C}_g$ in Eq.\ (\ref{e14}), and the relation between the gallery sample $g$ and the probe sample's context $\mathcal{C}_p$ in Eq.\ (\ref{e13}):
\begin{equation}\label{e15}
S_{p,g}=S_{p,\mathcal{C}_g}+S_{g,\mathcal{C}_p}.
\end{equation}
Compared to $C_p$ or $C_g$ as unilateral-context, both $C_p$ and $C_g$ as the bilateral-context are adopted and brings more effective information for computing $S_{p,g}$. Thus, better performance of person re-id can be obtained.

\subsubsection{Progressive optimization}
We propose a progressive optimization strategy to optimize the initial ranking list $\mathcal{R}^o_p$. Both first order context and second order context are utilized for the computation of pairwise similarity in sequence. In comparison to the first order context, the second order context is developed based on the high-order information and carry more reliable and effective contextual information. Therefore, we firstly 
adopt the second order context to refine $\mathcal{R}^o_p$ and approach the optimal solution of the problem. Then, we further fine-tune the results of the previous step by the first order context.

Specifically, we firstly compute the similarity between $p$ and $g$ with the aid of $\mathcal{C}^2(p,k_0,k)$ and $\mathcal{C}^2(g,k_0,k)$ such that
\begin{equation}\label{e11}
\widehat{S}_{p,g}=S_{p,\mathcal{C}^2_g}+S_{g,\mathcal{C}^2_p},
\end{equation}
where $\mathcal{C}^2_p=\mathcal{C}^2(p,k_0,k)$ and $\mathcal{C}^2_g=\mathcal{C}^2(g,k_0,k)$.
For each gallery sample from the collection of gallery samples, we compute the similarity with the probe sample $p$ by Eq.\ (\ref{e11}), respectively. And then an optimized ranking list of the probe sample $p$ is obtained. On this basis, we compute the first order context $\mathcal{C}^1(p,k)$, and leverage $\mathcal{C}^1(p,k)$ and $\mathcal{C}^1(g,k)$ to compute the final similarity between $p$ and $g$:
\begin{equation}\label{e12}
S_{p,g}=S_{p,\mathcal{C}^1_g}+S_{g,\mathcal{C}^1_p}, 
\end{equation}
where $\mathcal{C}^1_p=\mathcal{C}^1(p,k)$ and $\mathcal{C}^1_g=\mathcal{C}^1(g,k)$.

\begin{figure} 
\begin{center}
   \includegraphics[width=0.98\linewidth, height=0.7\linewidth]{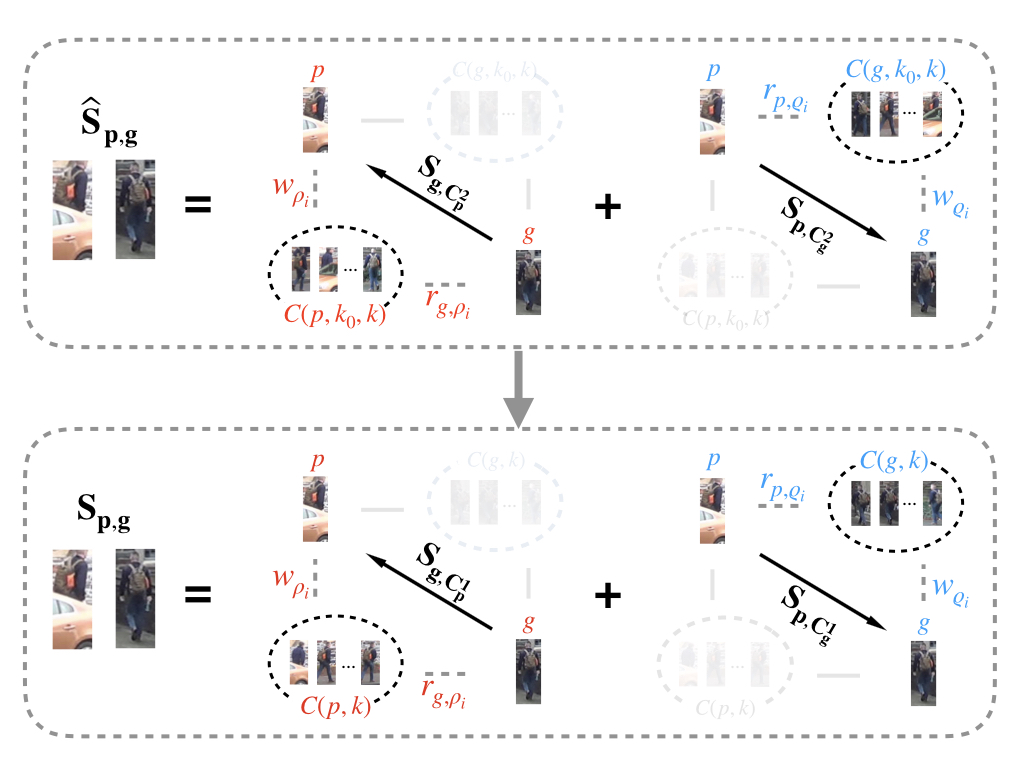}
\end{center}
   \caption{Overview of the proposed method. Better viewed in color.}
\label{fig4}
\end{figure}

Fig.~\ref{fig4} illustrates the computation procedure of the proposed method.  

\subsection{Efficiency}
\subsubsection{Efficiency enhancement}
\label{en}
The proposed method serves as a post-processing step after the content-based method to refine the initial ranking list. The efficiency of the method is one of important performance indexes. For the proposed method to solve the person re-id problem more efficiently, there are two technical improvements worth noting. (1) The number of gallery samples $N$ is usually very high in reality. It will be time consuming if we optimize the ranking positions of all gallery samples for a given probe sample. The top positions of the ranking list are expected to contain the most relevant samples
with regard to the probe sample, so we only re-compute the similarity between the probe sample and the gallery samples with top-$L$ positions of ranking list $\mathcal{R}^o_p$ ($L\ll N$). 
(2) The calculation related to the gallery sample can be done offline. For the calculation related to the probe sample, i.e.\ the similarity value $r_{p,\varrho_i}$ and the weight $w_{\rho_i}$, if we consider the ranking positions from both ranking lists in Eqs.\ (\ref{e6}) and (\ref{e7}), the algorithm will be time consuming, since they need to be computed online. Instead, we only consider the nonreciprocal rank information for these calculation like in Eq.\ (\ref{e3}). In doing so, the calculation related to the probe sample is done online with the simple retrieve operation, so that the calculation complexity will be significantly reduced.

\subsubsection{Complexity analysis}
In the proposed method, most operations, i.e.\ the calculation related to the gallery sample, can be done in advance offline with $O(N^2+NlogN)$ computation complexity. During the online phase, the computation complexity of the proposed method is $O(k_0kL+LlogL)$ for one probe sample. $L$ is far less than the number of gallery samples $N$. $k_0$ and $k$ are the parameters for computing the context and are very small values. By comparison, some post-processing person re-id methods \cite{zhong2017re,saquib2018pose} use the information of all probe and gallery samples for increasing the accuracy of person re-id. This will result in considerable computation complexity and space complexity. Moreover, it is 
not in accordance with the practical applications where only one probe sample is usually given. Detailed comparisons on run times will be presented in the next section.

\section{Experiments}
\label{ex}

\subsection{Datasets and settings}

\textbf{Datasets.}
We conduct the experiments on four large-scale person re-id benchmark datasets, including the \textbf{Market1501} dataset \cite{zheng2015scalable}, the \textbf{DukeMTMC} dataset \cite{ristani2016performance}, the \textbf{CUHK03} dataset \cite{li2014deepreid} and a video-based dataset \textbf{MARS} \cite{zheng2016mars}, in which there are multi person images for each of the gallery IDs, guaranteeing that the context of the sample can provide effective information for assisting the person re-id. In experiments, we adopt the standard splits of training and testing IDs for Market1501, DukeMTMC and MARS datasets as in \cite{suh2018part}. For CUHK03, we adopt the new training/testing protocol in \cite{zhong2017re}. Particularly, We report the experiment results on Market1501 dataset under the single-probe evaluation setting.

\textbf{Evaluation metrics.} Two evaluation metrics are adopted to evaluate the effectiveness of the proposed method: cumulated matching characteristics
(CMC), and mean average precision (mAP). Rank-k recognition rate in CMC is the expectation of finding the correct match within the first-k ranks and we report the results at Rank-1. The evaluation metric mAP considers both precision and recall. 

\subsection{Analysis of the proposed method}
In this section, we conduct the experiments on the Market1501 and CUHK03(detected) datasets with the content-based method IDE(R)+Euclidean \cite{zheng2017person} as the baseline, unless otherwise specified.

\subsubsection{Parameters Analysis} 
\begin{figure}
\begin{center}
   \includegraphics[width=0.85\linewidth, height=0.67\linewidth]{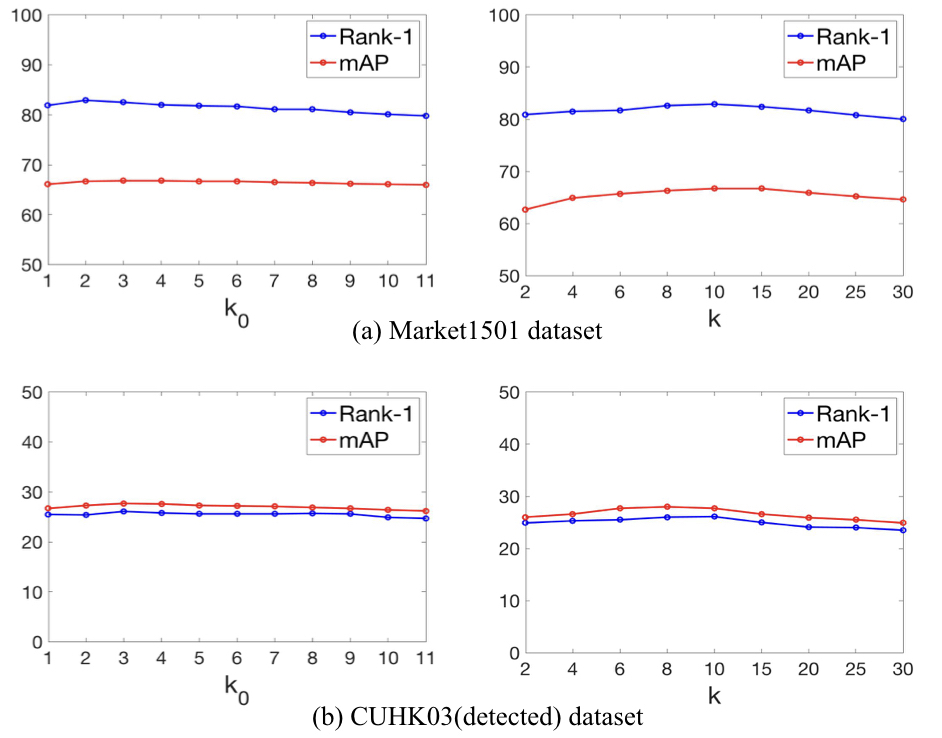}
\end{center}
   \caption{The impact of parameters $k_0$ and $k$ on the re-identification performance on the Market1501 and CUHK03(detected) datasets.}
\label{fig5}
\end{figure}

\begin{figure}
\begin{center}
   \includegraphics[width=0.9\linewidth, height=0.4\linewidth]{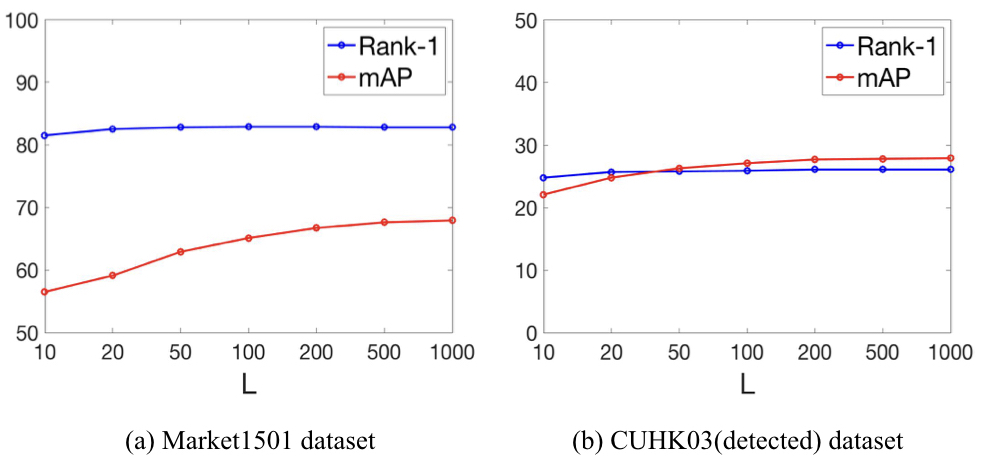}
\end{center}
   \caption{The impact of parameter $L$ on the re-identification performance on the Market1501 and CUHK03(detected) datasets.}
\label{fig7}
\end{figure}

There are two parameters in the proposed method: $k_0$ and $k$ for computing the target sample's context. In addition, considering the efficiency of algorithm, we only optimize the ranking positions of $L$ gallery samples instead of all gallery samples in the proposed method and the parameter $L$ needs to be determined. Therefore, in this subsection, we analyze in detail the impact of the parameters $k_0$, $k$ and $L$ on the performance of the person re-id. Fig.~\ref{fig5} and Fig.~\ref{fig7} show the impact of these parameters on the Rank-1 accuracy and the mAP. Each parameter is changed with other parameters setting as the default values. The default setting is $\left [ k_0\; k\; L \right ]=\left [ 2\; 10\; 200 \right ]$ for the Market1501 dataset and $\left [ k_0\; k\; L \right ]=\left [ 3\; 10\; 200 \right ]$ for the CUHK03(detected) dataset. \\
(1) As shown in Fig.~\ref{fig5}, with the increase of $k_0$ and $k$, the Rank-1 and mAP show slightly increasing at first and then slightly decreasing tendency. On the whole, the proposed method has good robustness to variation of parameters $k_0$ and $k$. 
When the parameters increases, more gallery samples similar to the target samples are introduced into the context and can bring a positive impact on improving the performance of the person re-id. However, the number of gallery samples having similar appearance with the target samples is limited. With the further increase of $k_0$ and $k$, interference samples will be introduced into the context to assist the person re-id. This results in a negative impact on the performance. 
Thanks to the introduction of the sample's weight in the context, even if the interference samples are introduced into the context with the increase of these two parameters, a small weight will likely be given for these samples and blunts the impact of these samples on the performance. As a result, there is a slight change on the performance with the increase of $k_0$ and $k$ in the proposed method.
Besides, we can also observe that the inflection point of the curve for $k_0$ appears at around $2-3$ and for $k$ at around $8-10$, which shows that the performance is more sensitive to the increase of $k_0$ than that of $k$. This is reasonable, because we can see from the definition of the context that the top-$k_0$ samples have more influence on the quality of the context. Therefore, a small $k_0$ should be given and we set $k_0=2$ for the Market1501, DukeMTMC and MARS datasets, and $k_0=3$ for CUHK03 dataset in our experiments; and a slightly larger value of $k$ is given and we set $k=10$ for all datasets in our experiments.\\
(2) As shown in Fig.~\ref{fig7}, with the increase of $L$, the Rank-1 gradually increases and finally tends to be stable, while the mAP continues to increase with rapid increase rate in the beginning and gradually slow afterwards. This is because the increase of $L$ means that more gallery samples with low ranking in the initial ranking list will participate in the optimization of ranking position by the proposed method. There are differences on appearance between these gallery samples and the probe sample, so that the probe sample's (these gallery samples') context is also not similar to these gallery samples (probe sample) and the contextual information can not assist the person re-id, resulting in that the ranking positions of these gallery samples can not be optimized effectively by the proposed method. By comprehensive consideration of CMC, mAP and the efficiency of method, we set $L=200$ for all datasets in experiments.

\begin{table}
\begin{center}
\caption{Comparison of the performance for the proposed method with different contexts on the Market1501 and CUHK03(detected) datasets. The best results are shown in red/bold. Better viewed in color.}\label{tab1}
\renewcommand\arraystretch{1.0}
\begin{tabular}{l||p{0.9cm}|p{0.7cm}||p{0.9cm}|p{0.7cm}}
\Xhline{1.2pt}
\multirow{2}{*}{Methods}& \multicolumn{2}{c||}{Market1501} & \multicolumn{2}{c}{CUHK03(detected)} \\
\cline{2-5}
 & Rank-1 & mAP & Rank-1 & mAP \\
\hline
IDE(R)+Euclidean(baseline) & 78.9 & 55.0 & 21.4 & 19.8 \\
\hline
Ours$_p$ & 80.9 & 64.4 & 25.1 & 25.8 \\
Ours$_g$ & 81.8 & 65.3 & 25.0 & 26.1 \\
\hline
Ours$_f$ & 81.1 & 63.9 & 24.8 & 25.3 \\
Ours$_s$ & 82.2 & 66.6 & 25.8 & 27.4 \\
\hline
Ours & \textcolor[rgb]{1,0,0}{\textbf{82.9}} & \textcolor[rgb]{1,0,0}{\textbf{66.7}} & \textcolor[rgb]{1,0,0}{\textbf{26.1}} & \textcolor[rgb]{1,0,0}{\textbf{27.7}} \\
\Xhline{1.2pt}
\end{tabular}
\end{center}
\end{table}

\subsubsection{Analysis on context set} We propose a progressive bilateral-context driven optimization strategy. The second order context and the first order context are utilized to assist the person re-id in sequence, and both contexts of probe sample and gallery sample are adopted at each stage. In this subsection, we investigate the effects of these contexts on the performance.
Table~\ref{tab1} reports the performance comparison of the proposed method with different contexts. 
We can see that 
(1) whatever we adopt the probe sample's context (Ours$_p$) or the gallery sample's context (Ours$_g$) as unilateral-context to assist person re-id, the best performance is obtained by the proposed method with effectively combining these two contexts as bilateral-context;
(2) in general, the proposed method with the gallery sample's context is superior to the one with the probe sample's context on performance, which means that the context of the gallery sample is more helpful to refine the initial results;
(3) the proposed method with progressive optimization strategy yields more satisfying results, compared to only utilizing the first order context (Ours$_f$) or the second order context (Ours$_s$);
(4) compared with the first order context, the second order context has higher positive effect on the performance, which reflects that more valuable information exists in the second k-nn context.

\subsubsection{Analysis on the similarity between target sample and sample in counterpart’s context} In Section~\ref{csv}, we analyzed in detail the existing problem of using nonreciprocal rank information in Eq.\ (\ref{e3}), the reciprocal rank-based measure in Eq.\ (\ref{e4}) or in Eq.\ (\ref{e5}) for computing the similarity between the target sample and the sample in counterpart’s context. We drew the conclusion that the similarity can be more accurately computed by combing the reciprocal rank-based measure Eq.\ (\ref{e4}) and Eq.\ (\ref{e5}). In this subsection, we evaluate the performance of the proposed method with different similarity measures to validate our conclusion. We report the comparison results in Table~\ref{tab2}. The proposed method based on the nonreciprocal rank information (Our$_{n}$), or with any one of either the reciprocal rank-based measure in Eq.\ (\ref{e4}) (Our$_{r1}$) or in Eq.\ (\ref{e5}) (Our$_{r2}$) has a limited effect on the performance and even causes the decline of the performance in some cases. For example, Ours$_{r1}$ reduces the mAP of the baseline from 55.0\% to 41.5\% on the Market1501 dataset. However, by adopting the reciprocal rank information with the combination of these two measures, the proposed method consistently improves the Rank-1 accuracy and the mAP of the baseline method, and outperforms the proposed method with any one of reciprocal rank-based measures and the one with the nonreciprocal rank information.

\begin{table}
\begin{center}
\caption{Comparison of the performance for the proposed method with different measures of similarity between sample and context on the Market1501 and CUHK03(detected) datasets. The best results are shown in red/bold. Better viewed in color.}\label{tab2}
\renewcommand\arraystretch{1.0}
\begin{tabular}{l||p{0.9cm}|p{0.7cm}||p{0.9cm}|p{0.7cm}}
\Xhline{1.2pt}
\multirow{2}{*}{Methods}& \multicolumn{2}{c||}{Market1501} & \multicolumn{2}{c}{CUHK03(detected)} \\
\cline{2-5}
 & Rank-1 & mAP & Rank-1 & mAP \\
\hline
IDE(R)+Euclidean(baseline) & 78.9 & 55.0 & 21.4 & 19.8 \\
\hline
Ours$_{n}$ & 80.5 & 65.9 & 25.9 & 27.5 \\
Ours$_{r1}$ & 79.8 & 41.5 & 24.6 & 21.0 \\
Ours$_{r2}$ & 82.8 & 66.5 & 25.7 & 27.6 \\
\hline
Ours & \textcolor[rgb]{1,0,0}{\textbf{82.9}} & \textcolor[rgb]{1,0,0}{\textbf{66.7}} & \textcolor[rgb]{1,0,0}{\textbf{26.1}} & \textcolor[rgb]{1,0,0}{\textbf{27.7}} \\
\Xhline{1.2pt}
\end{tabular}
\end{center}
\end{table}

\begin{table}
\begin{center}
\caption{Comparison of the performance for the proposed method with different versions for the weight of the sample in the context on the Market1501 and CUHK03(detected) datasets. The best results are shown in red/bold. Better viewed in color.}\label{tab3}
\renewcommand\arraystretch{1.0}
\begin{tabular}{l||p{0.9cm}|p{0.7cm}||p{0.9cm}|p{0.7cm}}
\Xhline{1.2pt}
\multirow{2}{*}{Methods}& \multicolumn{2}{c||}{Market1501} & \multicolumn{2}{c}{CUHK03(detected)} \\
\cline{2-5}
 & Rank-1 & mAP & Rank-1 & mAP \\
\hline
IDE(R)+Euclidean(baseline) & 78.9 & 55.0 & 21.4 & 19.8 \\
\hline
Ours(\emph{unweighted}) & 80.6 & 65.5 & 25.4 & 26.0 \\
Ours(\emph{no-reliability}) & 82.7 & 66.6 & 26.1 & \textcolor[rgb]{1,0,0}{\textbf{27.8}} \\
\hline
Ours & \textcolor[rgb]{1,0,0}{\textbf{82.9}} & \textcolor[rgb]{1,0,0}{\textbf{66.7}} & \textcolor[rgb]{1,0,0}{\textbf{26.1}} & 27.7 \\
\Xhline{1.2pt}
\end{tabular}
\end{center}
\end{table}

\begin{table*}[ht]
\begin{center}
\caption{Comparison of the performance for the proposed method with various baseline methods on the Market1501 and CUHK03(detected) datasets. 'Offline' denotes the run time of offline for all gallery samples. 'Online-Single' and 'Online-Total' denote the run time of online for one probe sample and all probe samples, respectively.}\label{tab4}
\renewcommand\arraystretch{1.0}
\begin{tabular}{p{3.1cm}<{\centering}||p{1.4cm}<{\centering}|p{1.5cm}<{\centering}|p{0.7cm}<{\centering}|p{0.8cm}<{\centering}|p{0.7cm}<{\centering}||p{1.5cm}<{\centering}|p{1.5cm}<{\centering}|p{0.7cm}<{\centering}|p{0.8cm}<{\centering}|p{0.7cm}<{\centering}}
\Xhline{1.2pt}
\multirow{4}{*}{Methods} & \multicolumn{5}{c||}{Market1501} & \multicolumn{5}{c}{CUHK03(detected)} \\
\cline{2-11}
 & \multirow{3}{*}{Rank-1} & \multirow{3}{*}{mAP} & \multicolumn{3}{c||}{Time(s)} & \multirow{3}{*}{Rank-1} & \multirow{3}{*}{mAP} & \multicolumn{3}{c}{Time(s)} \\
 \cline{4-6}  \cline{9-11}
 & & & \multirow{2}{*}{Offline} & \multicolumn{2}{c||}{Online} & & & \multirow{2}{*}{Offline} & \multicolumn{2}{c}{Online}\\
  \cline{5-6}  \cline{10-11}
 & & & & Single & Total & & & & Single & Total \\
\hline
LOMO+XQDA(baseline) & 47.8 & 24.7 & ~~- & ~~- & ~~- & 12.3 & 11.5 & ~~- & ~~- & ~~-\\
Ours & 54.7 ($\uparrow$6.9) & 34.5 ($\uparrow$9.8) & 22.2 & 6.6e-3 & 22.1 & 15.9 ($\uparrow$3.6) & 17.2 ($\uparrow$5.7) & 2.0 & 4.1e-3 & 5.7\\
\hline
IDE(R)+XQDA(baseline) & 78.0 & 56.2 & ~~- & ~~- & ~~- & 30.7 & 28.5 & ~~- & ~~- & ~~- \\
Ours & 81.7 ($\uparrow$3.7) & 66.5 ($\uparrow$10.3) & 23.5 & 6.5e-3 & 22.0 & 37.9 ($\uparrow$7.2) & 38.7 ($\uparrow$10.2) & 1.8 & 4.5e-3 & 6.3 \\
\hline
PCB(baseline) & 93.1 & 80.4 & ~~- & ~~- & ~~- & 54.9 & 50.7 & ~~- & ~~- & ~~- \\
Ours & 94.2 ($\uparrow$1.1) & 88.9 ($\uparrow$8.5) & 26.5 & 5.6e-3 & 18.8 & 65.9 ($\uparrow$11.0) & 66.1 ($\uparrow$15.4) & 1.8 & 4.7e-3 & 6.6 \\
\Xhline{1.2pt}
\end{tabular}
\end{center}
\end{table*}

\begin{table*}[ht]
\begin{center}
\caption{Comparison with the proposed method based on the similarity value on the Market1501 and CUHK03(detected) datasets. }\label{tab5}
\renewcommand\arraystretch{1.0}
\begin{tabular}{l||p{1.5cm}|p{1.5cm}||p{1.5cm}|p{1.5cm}}
\Xhline{1.2pt}
\multirow{2}{*}{Methods}& \multicolumn{2}{c||}{Market1501} & \multicolumn{2}{c}{CUHK03(detected)} \\
\cline{2-5}
 & Rank-1 & mAP & Rank-1 & mAP \\
\hline
IDE(R)+Euclidean(baseline) & 78.9 & 55.0 & 21.4 & 19.8 \\
Ours$_v$ & 69.7 ($\downarrow$9.2) & 56.4 ($\uparrow$1.4) & 19.4 ($\downarrow$2.0) & 21.0 ($\uparrow$1.2) \\
Ours & 82.9 ($\uparrow$4.0) & 66.7 ($\uparrow$11.7) & 26.1 ($\uparrow$4.7) & 27.7 ($\uparrow$7.9) \\
\hline
IDE(R)+XQDA(baseline) & 78.0 & 56.2 & 30.7 & 28.5 \\
Ours$_v$ & 79.0 ($\uparrow$1.0) & 61.5 ($\uparrow$5.3) & 33.6 ($\uparrow$2.9) & 33.7 ($\uparrow$5.2) \\
Ours & 81.7 ($\uparrow$3.7) & 66.5 ($\uparrow$10.3) & 37.9 ($\uparrow$7.2) & 38.7 ($\uparrow$10.2) \\
\Xhline{1.2pt}
\end{tabular}
\end{center}
\end{table*}

\subsubsection{Analysis on the weight of the sample in the context} We deem that the sample in the context with more similarity to the target sample is more reliable to assist the person re-id, so the samples in context are given different weights according to the similarity with the target sample in Section~\ref{wsc}. In this subsection, we validate the effectiveness of the proposed method with weighted context. In addition, we introduced the ranking list's reliability to the computation of the weight, and its effectiveness is evaluated in this subsection. The comparison results are shown in Table~\ref{tab3}. We observe the following:
(1) When the context samples are weighted equally and are used to assist person re-id in the proposed method (Ours(\emph{unweighted})), the performance of the baseline method can be boosted. However, when different weights are given to the context samples, the proposed method has superior performance on both datasets. 
(2) Compared with the proposed method in which the ranking list's reliability is not introduced (Ours(\emph{no-reliability})), a better performance is achieved by introducing the reliability into the computation of the weight in the proposed method on the Market1501 dataset. However, the proposed method with or without the reliability is almost same in terms of performance on the CUHK03(detected) dataset. It might be because a baseline method with low Rank-1 and mAP is given on the CUHK03(detected) dataset and we compute the ranking list's reliability based on these initial results, resulting in an inaccurate measure of reliability and the limited effect on the performance. 

\subsubsection{Effectiveness, high efficiency and robustness} 
\begin{figure}
\begin{center}
   \includegraphics[width=1\linewidth, height=0.45\linewidth]{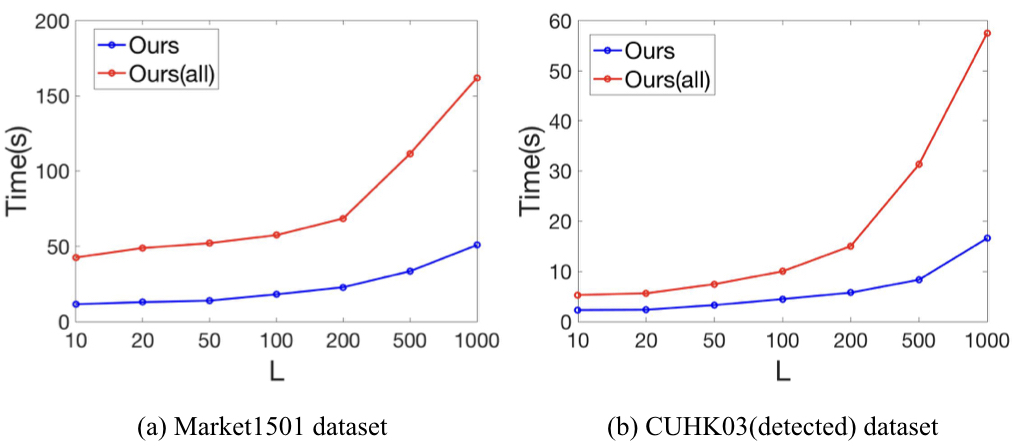}
\end{center}
   \caption{Comparison with different versions for the proposed method on run time on the Market1501 and CUHK03(detected) datasets.}
\label{fig10}
\end{figure}

In this subsection, we validate the the advantages of the proposed method:
effectiveness, high efficiency and robustness.\\
(1) Effectiveness. We adopt the results from three different content-based methods LOMO+XQDA \cite{liao2015person}, IDE(R)+XQDA \cite{zheng2017person} and PCB \cite{sun2018beyond} as the input of the proposed method, respectively. The experimental results are shown in Table~\ref{tab4}. We can see that the proposed method can effectively improve the Rank-1 and mAP with different content-based methods as the baselines, showing the effectiveness of the proposed method. In the next section, we will further verify the effectiveness of the proposed method compared to the state-of-the-art methods. \\
(2) High efficiency\footnote{The experiments are conducted on a server with Intel Xeon CPU (2.6GHz) and 128GB RAM.}. From Table~\ref{tab4}, it can also be seen that during the offline phase, the run time of proposed method is at most $26.5$ seconds on the Market1501 dataset with $15,913$ gallery samples and $2.0$ seconds on the CUHK03(detected) dataset with $5,332$ gallery samples, during the online phase, the run time is at most $22.1$ seconds on the Market1501 dataset with $3,368$ queries, and $6.6$ seconds on the CUHK03(detected) dataset with $1,400$ queries, it takes an average of $5.3$ milliseconds to optimize the initial ranking results of one probe sample by the proposed method, which fully reflects the high efficiency of the proposed method. 
Its high efficiency is creditable to the top ranked sample optimization and the gallery based offline computation, which is verified by Fig.~\ref{fig10}. We can see that, 
(1) with the increase of the number of gallery samples to be re-ranked ($L$), the run time of the proposed method (Ours) increases continuously, we set $L=200$ in experiments with consideration of the efficiency;
(2) compared to the proposed method in which all probe and gallery samples are used to assist re-id and results in all sample based offline computation (Ours(all)), the proposed method with gallery samples used to assist re-id and the gallery based offline computation (Ours) can achieve faster re-ranking speed.
We will see that the proposed method still have advantage on efficiency compared to the state-of-the-art methods in the next section.\\
(3) Robustness. All calculation are based on the rank information in the proposed method, resulting in its robustness to the variation of the similarity value from the baseline method. For validating the robustness, we also report the results of the proposed method that optimizes the initial results based on the similarity values of baseline method (Ours$_v$) in Table~\ref{tab5}. When we adopt two different baseline methods, meaning the variation of the similarity value, the performance enhancement can not always be achieved by Ours$_v$. For example, the Rank-1 accuracy drops $9.2\%$ and $2.0\%$ on the Market1501 and CUHK03(detected) datasets with IDE(R)+Euclidean baseline method, respectively. In contrast, the proposed method based on the rank information can consistently enhance the performance with these two different baselines, indicating its robustness to variation of the similarity value. 

\subsection{Comparison with state-of-the-art}
In this section, we compare the performance of the proposed method with the published non-post-processing and post-processing person re-id methods on the Market1501, DukeMTMC, CUHK03 and MARS datasets.
For a fair comparison, we adopt the same content-based method as the baseline of all post-processing methods in experiments.
Particularly, for the post-processing method, we consider that the run time of the algorithm is also one of the key performance indexes like the Rank-1 and mAP metrics. Therefore, we compare the proposed method with other state-of-the-art post-processing methods on not only Rank-1 and mAP, but also the run time of the algorithm.

\begin{table}
\begin{center}
\caption{Comparison with various state-of-the-art methods on the Market1501 dataset. Time(s) indicates the run time in seconds of the method. In the comparison among post-processing methods, the best and second results are shown in red/bold and blue/bold, respectively. Better viewed in color.}\label{tab6}
\renewcommand\arraystretch{1.0}
\begin{tabular}{l|l||p{0.9cm}|p{0.9cm}|p{0.7cm}}
\Xhline{1.2pt}
 \multicolumn{2}{l||}{\textbf{Methods}} & Rank-1 & mAP & Time(s) \\
\hline
 Gated SCNN \cite{varior2016gated} & ECCV2016 & 65.9 & 39.6 & ~~-\\
 PDC \cite{su2017pose} & ICCV2017 & 84.1 & 63.4 & ~~-\\
 LSRO \cite{zheng2017unlabeled} & ICCV2017 & 84.0 & 66.1 & ~~-\\
 LML(S2S) \cite{zhou2018large} & TMM2018 & 65.3 & 39.8 & ~~-\\
 MVLDML+ \cite{yang2018person} & TIP2018 & 58.2 & 33.7 & ~~-\\
 HA-CNN \cite{li2018harmonious} & CVPR2018 & 91.2 & 75.7 & ~~- \\
 DGSRW \cite{shen2018deep} & CVPR2018 & 92.7 & 82.5 & ~~-\\
 GLIA \cite{chen2018improving} & ECCV2018 & 93.3 & 81.8 & ~~-\\
 SGGNN \cite{shen2018person} & ECCV2018 & 92.3 & 82.8 & ~~-\\
 $P^2$-Net(+triplet loss) \cite{Guo_2019_ICCV} & ICCV2019 & 95.2 & 85.6 & ~~-\\
 BDB+Cut \cite{dai2019batch} & ICCV2019 & 95.3 & 86.7 & ~~-\\
 RFD \cite{Yang_2019_CVPR} & CVPR2019 & 94.7 & 84.5 & ~~-\\
 Pyramid \cite{Zheng_2019_CVPR} & CVPR2019 & 95.7 & 88.2 & ~~-\\
\hline
 BTricks(baseline) \cite{luo2019bag} & CVPR2019 & 94.3 & 85.4 & ~~-\\
 k-reciprocal \cite{zhong2017re} & CVPR2017 & \textcolor[rgb]{0,0,1}{\textbf{95.2}} & \textcolor[rgb]{1,0,0}{\textbf{93.9}} & \textcolor[rgb]{0,0,1}{\textbf{74.5}} \\
 ECN(orig-dist) \cite{saquib2018pose} & CVPR2018 & 94.6 & 89.9 & 74.9  \\
 ECN(rank-dist) \cite{saquib2018pose} & CVPR2018 & 94.6 & 92.1 & 117.4 \\
 Ours & & \textcolor[rgb]{1,0,0}{\textbf{95.6}} & \textcolor[rgb]{0,0,1}{\textbf{92.2}} & \textcolor[rgb]{1,0,0}{\textbf{20.6}} \\
\Xhline{1.2pt}
\end{tabular}
\end{center}
\end{table}

\subsubsection{Market1501}
In Table~\ref{tab6}, we report the comparison results with $13$ non-post-processing methods and $3$ post-processing methods. 
Compared with the non-post-processing methods, the proposed method surpasses all methods on mAP, and gets the second best result at Rank-1 and performs slightly worse than the Pyramid \cite{Zheng_2019_CVPR} by $0.1\%$ at Rank-1. However, the proposed method is very likely to have a better performance with a better baseline.
Compared with the post-processing methods, the proposed method clearly outperforms other methods at Rank-1 and yields the second best performance on mAP. For the run time of the algorithm, the proposed method is almost $4-7$ times faster than the k-reciprocal \cite{zhong2017re} and ECN \cite{saquib2018pose} methods.
In addition, it is worth mentioning that the proposed method outperforms the context-based deep learning methods DGSRW \cite{shen2018deep} and SGGNN \cite{shen2018person} by a large margin. DGSRW and SGGNN aim to learn an optimal feature representation with the aid of the contextual information and are the end-to-end deep learning framework. In comparison, the proposed method aims to enhance the performance of person re-id by utilizing the contextual information and is a post-processing method.

\subsubsection{DukeMTMC}

We compare the proposed method with other state-of-the-art person re-id methods on the DukeMTMC dataset, as listed in Table~\ref{tab7}. The proposed method clearly outperforms other existing non-post-processing methods. In addition, for the comparison among post-processing methods, the proposed method yields the second best performance at Rank-1. 
However, the proposed method has an absolute advantage on run time. The ECN(orig-dist) \cite{saquib2018pose} method executes in $45.4$ seconds to achieve the best performance of $91.4\%$ at Rank-1. In contrast, we only need $14.4$ seconds in the proposed method and obtain $91.2\%$ at Rank-1, nearly equivalent to the Rank-1 result of the ECN(orig-dist) method while using less than one third of the run time.
Besides, compared to the context-based deep learning methods DGSRW \cite{shen2018deep} and SGGNN \cite{shen2018person}, the proposed method still have an overwhelming advantage in performance.
 
\begin{table}
\begin{center}
\caption{Comparison with various state-of-the-art methods on the DukeMTMC dataset. In the comparison among post-processing methods, the best and second results are shown in red/bold and blue/bold, respectively. Better viewed in color.}\label{tab7}
\renewcommand\arraystretch{1.0}
\begin{tabular}{l|l||p{0.9cm}|p{0.9cm}|p{0.7cm}}
\Xhline{1.2pt}
 \multicolumn{2}{l||}{\textbf{Methods}} & Rank-1 & mAP & Time(s) \\
\hline
 LSRO \cite{zheng2017unlabeled} & ICCV2017 & 67.7 & 47.1 & ~~- \\
 SVDNet \cite{sun2017svdnet} & ICCV2017 & 76.7 & 56.8 & ~~- \\
 HA-CNN \cite{li2018harmonious} & CVPR2018 & 80.5 & 63.8 & ~~- \\
 SPReID \cite{kalayeh2018human} & CVPR2018 & 82.0 & 73.3 & ~~- \\
 DGSRW \cite{shen2018deep} & CVPR2018 & 80.7 & 66.4 & ~~-\\
 PABR \cite{suh2018part} & ECCV2018 & 82.1 & 64.2 & ~~- \\
 Mancs \cite{wang2018mancs} & ECCV2018 & 84.9 & 71.8 & ~~- \\
 SGGNN \cite{shen2018person} & ECCV2018 & 81.1 & 68.2 & ~~-\\
 $P^2$-Net(+triplet loss) \cite{Guo_2019_ICCV} & ICCV2019 & 86.5 & 73.1 & ~~-\\
 BDB+Cut \cite{dai2019batch} & ICCV2019 & 89.0 & 76.0 & ~~-\\
 RFD \cite{Yang_2019_CVPR} & CVPR2019 & 85.8 & 72.9 & ~~-\\
 Pyramid \cite{Zheng_2019_CVPR} & CVPR2019 & 89.0 & 79.0 & ~~-\\
\hline
 BTricks(baseline) \cite{luo2019bag} & CVPR2019 & 86.0 & 75.1 & ~~-\\
 k-reciprocal \cite{zhong2017re} & CVPR2017 & 90.8 & \textcolor[rgb]{0,0,1}{\textbf{88.6}} & 75.3 \\
 ECN(orig-dist) \cite{saquib2018pose} & CVPR2018 & \textcolor[rgb]{1,0,0}{\textbf{91.4}} & 86.5 & \textcolor[rgb]{0,0,1}{\textbf{45.4}}  \\
 ECN(rank-dist) \cite{saquib2018pose} & CVPR2018 & 91.0 & \textcolor[rgb]{1,0,0}{\textbf{89.2}} & 75.1 \\
 Ours & & \textcolor[rgb]{0,0,1}{\textbf{91.2}} & 84.7 & \textcolor[rgb]{1,0,0}{\textbf{14.4}} \\
\Xhline{1.2pt}
\end{tabular}
\end{center}
\end{table}

\begin{table}
\begin{center}
\caption{Comparison of the proposed method with post-processing methods on the MARS dataset. The best and second results are shown in red/bold and blue/bold. Better viewed in color.}\label{tab9}
\renewcommand\arraystretch{1.0}
\begin{tabular}{l||p{0.9cm}|p{0.9cm}|p{0.7cm}}
\Xhline{1.2pt}
 \textbf{Methods} & Rank-1 & mAP & Time(s) \\
\hline
 IDE(C)+Euclidean(baseline) \cite{zheng2017person} & 60.8 & 41.2 & ~~-\\
 k-reciprocal \cite{zhong2017re} & 63.6 & \textcolor[rgb]{0,0,1}{\textbf{51.9}} & 26.2 \\
 ECN(orig-dist) \cite{saquib2018pose} & 64.4 & 49.9 & \textcolor[rgb]{0,0,1}{\textbf{21.0}} \\
 ECN(rank-dist) \cite{saquib2018pose} & \textcolor[rgb]{0,0,1}{\textbf{65.0}} & \textcolor[rgb]{1,0,0}{\textbf{53.0}} & 30.7 \\
 Ours & \textcolor[rgb]{1,0,0}{\textbf{66.7}} & 51.5 & \textcolor[rgb]{1,0,0}{\textbf{11.7}} \\
 \hline
 IDE(C)+KISSME(baseline) \cite{zheng2017person} & 65.0 & 44.2 & ~~-\\
 k-reciprocal \cite{zhong2017re} & 66.8 & \textcolor[rgb]{0,0,1}{\textbf{57.3}} & 27.8 \\
 ECN(orig-dist) \cite{saquib2018pose} & \textcolor[rgb]{0,0,1}{\textbf{68.1}} & 54.1 & \textcolor[rgb]{0,0,1}{\textbf{20.6}} \\
 ECN(rank-dist) \cite{saquib2018pose} & 67.9 & \textcolor[rgb]{1,0,0}{\textbf{57.6}} & 31.3 \\
 Ours & \textcolor[rgb]{1,0,0}{\textbf{69.2}} & 55.1 & \textcolor[rgb]{1,0,0}{\textbf{10.6}} \\
 \hline
 IDE(C)+XQDA(baseline) & 65.5 & 46.9 & ~~-\\
 k-reciprocal \cite{zhong2017re} & 68.4 & \textcolor[rgb]{0,0,1}{\textbf{58.5}} & 29.3 \\
 ECN(orig-dist) \cite{saquib2018pose} & 68.1 & 55.8 & \textcolor[rgb]{0,0,1}{\textbf{20.7}} \\
 ECN(rank-dist) \cite{saquib2018pose} & \textcolor[rgb]{0,0,1}{\textbf{69.0}} & \textcolor[rgb]{1,0,0}{\textbf{59.0}} & 31.3 \\
 Ours & \textcolor[rgb]{1,0,0}{\textbf{69.0}} & 57.2 & \textcolor[rgb]{1,0,0}{\textbf{10.3}} \\
\Xhline{1.2pt}
\end{tabular}
\end{center}
\end{table}

\begin{table*}
\begin{center}
\caption{Comparison with various state-of-the-art methods on the CUHK03 dataset. $^*$ denotes an unpublished paper. In the comparison among post-processing methods, the best and second results are shown in red/bold and blue/bold. Better viewed in color.}\label{tab8}
\renewcommand\arraystretch{1.0}
\begin{tabular}{l|l||p{1cm}|p{1cm}|p{1cm}|p{1cm}|p{1cm}|p{1cm}}
\Xhline{1.2pt}
 \multicolumn{2}{l||}{\multirow{2}{*}{\textbf{Methods}}} & \multicolumn{3}{c|}{labeled} & \multicolumn{3}{c}{detected} \\
 \cline{3-8}
 \multicolumn{2}{l||}{} & Rank-1 & mAP & Time(s) & Rank-1 & mAP & Time(s) \\
 \hline
 TriNet+REDA$^*$ \cite{zhong2017random} & arXiv2017 & 58.1 & 53.8 & ~~- & 55.5 & 50.7 & ~~- \\
 SVDNet \cite{sun2017svdnet} & ICCV2017 & 40.9 & 37.8 & ~~- & 41.5 & 37.3 & ~~- \\
 HA-CNN \cite{li2018harmonious} & CVPR2018 & 44.4 & 41.0 & ~~- & 41.7 & 38.6 & ~~- \\
 Pose-Transfer \cite{liu2018pose} & CVPR2018 & 45.1 & 42.0 & ~~- & 41.6 & 38.7 & ~~- \\
 Mancs \cite{wang2018mancs} & ECCV2018 & 69.0 & 63.9 & ~~- & 65.5 & 60.5 & ~~- \\
 $P^2$-Net(+triplet loss) \cite{Guo_2019_ICCV} & ICCV2019 & 78.3 & 73.6 & ~~- & 74.9 & 68.9 & ~~- \\
 BDB+Cut \cite{dai2019batch} & ICCV2019 & 79.4 & 76.7 & ~~- & 76.4 & 73.5 & ~~- \\
 RFD \cite{Yang_2019_CVPR} & CVPR2019 & 70.1 & 66.5 & ~~- & 66.6 & 64.2 & ~~- \\
 Pyramid \cite{Zheng_2019_CVPR} & CVPR2019 & 78.9 & 76.9 & ~~- & 78.9 & 74.8 & ~~- \\
 \hline
  MHN-6(PCB)(baseline) \cite{Chen_2019_ICCV} & ICCV2019 & 75.7 & 70.5 & ~~- & 71.4 & 65.6 & ~~- \\
 k-reciprocal \cite{zhong2017re} & CVPR2017 & 82.2 & 82.0 & 8.4 & 77.7 & 77.4 & 8.0 \\
 ECN(orig-dist) \cite{saquib2018pose} & CVPR2018 & 84.1 & 84.0 & \textcolor[rgb]{0,0,1}{\textbf{5.3}} & \textcolor[rgb]{0,0,1}{\textbf{80.9}} & \textcolor[rgb]{0,0,1}{\textbf{80.4}} & \textcolor[rgb]{0,0,1}{\textbf{5.5}} \\
 ECN(rank-dist) \cite{saquib2018pose} & CVPR2018 & \textcolor[rgb]{0,0,1}{\textbf{84.1}} & \textcolor[rgb]{1,0,0}{\textbf{84.8}} & 7.7 & \textcolor[rgb]{1,0,0}{\textbf{81.3}} & \textcolor[rgb]{1,0,0}{\textbf{81.4}} & 7.6 \\
 Ours & & \textcolor[rgb]{1,0,0}{\textbf{84.4}} & \textcolor[rgb]{0,0,1}{\textbf{84.0}} & \textcolor[rgb]{1,0,0}{\textbf{5.1}} & 80.0 & 79.4 & \textcolor[rgb]{1,0,0}{\textbf{5.3}} \\
\Xhline{1.2pt}
\end{tabular}
\end{center}
\end{table*}

\subsubsection{CUHK03}
We conduct experiments on both CUHK03(labeled) and CUHK03(detected) datasets with comparison to several state-of-the-art methods. The comparison results are presented in Table~\ref{tab8}. For the 'labeled' data on the CUHK03 dataset, the proposed method achieves the best performance on Rank-1 and run time and obtains the second best performance on mAP in all comparison. The proposed method performs slightly worse than the ECN(rank-dist) \cite{saquib2018pose} method by $0.8\%$ on mAP. For the 'detected' data on CUHK03 dataset, the proposed method is only inferior to the state-of-the-art performances of ECN(rank-dist) by $1.3\%$ at Rank-1 and $2.0\%$ on mAP. However, the proposed method required less computation times than the other post-processing methods. 

\subsubsection{MARS}
We also compare the proposed method with other post-processing methods on a video-based dataset MARS. The IDE(C) feature descriptor is obtained by training a deep learning network based on the CaffeNet \cite{NIPS2012_4824} and is used as the representation for each frame. We aggregate the representations of all the consecutive frames using temporal average pooling. As a result, there are totally $1,980$ feature representation for the probe samples and $12,180$ feature representation for the gallery samples.
The IDE(C)+Euclidean \cite{zheng2017person}, IDE(C)+KISSME \cite{zheng2017person} and IDE(C)+XQDA are adopted as the baseline of the post-processing methods, respectively. The results are shown in Table~\ref{tab9}. As we can see, the proposed method consistently achieves the best performance on Rank-1 and run time in all comparison with different baselines. 

\begin{table*}
\begin{center}
\caption{Comparison with ECN method on the Market1501 and DukeMTMC datasets. 'Time(s)-PC' and 'Time(s)-Server' denote the run time on a PC ($3.4$GHz Intel Core CPU, $32$GB RAM) and a server($2.6$GHz Intel Xeon CPU, $128$GB RAM), respectively.}\label{tab10}
\renewcommand\arraystretch{1.0}
\begin{tabular}{l||p{0.9cm}<{\centering}|p{0.9cm}<{\centering}|p{0.9cm}<{\centering}|p{0.9cm}<{\centering}||p{0.9cm}<{\centering}|p{0.9cm}<{\centering}|p{0.9cm}<{\centering}|p{0.9cm}<{\centering}}
\Xhline{1.2pt}
\multirow{3}{*}{Methods}& \multicolumn{4}{c||}{Market1501} & \multicolumn{4}{c}{DukeMTMC} \\
\cline{2-9}
 & \multirow{2}{*}{Rank-1} & \multirow{2}{*}{mAP} & \multicolumn{2}{c||}{Time(s)} & \multirow{2}{*}{Rank-1} & \multirow{2}{*}{mAP} & \multicolumn{2}{c}{Time(s)} \\
 \cline{4-5} \cline{8-9}
 & & & PC & Server & & & PC & Server \\
\hline
PCB(baseline) \cite{sun2018beyond} & 93.1 & 80.4 & ~~- & ~~- & 85.0 & 72.4 & ~~- & ~~- \\
ECN(orig-dist) \cite{saquib2018pose} & 94.4 & 87.7 & 18970.7 & 74.1 & 88.8 & 81.0 & 1471.9 & 45.0\\
ECN(rank-dist) \cite{saquib2018pose} & 94.6 & 92.1 & 13926.2 & 111.3 & 90.2 & 87.8 & 2426.6 & 73.4 \\
Ours & 94.2 & 88.9 & 9.3 & 18.8 & 90.4 & 83.0 & 5.6 & 12.5 \\
\Xhline{1.2pt}
\end{tabular}
\end{center}
\end{table*}

It is worth mentioning that there are two main differences between the proposed method and the above these compared post-processing methods k-reciprocal \cite{zhong2017re} and ECN \cite{saquib2018pose}: (1) these two methods leverage the expended information of all samples in probe set in addition to gallery set while we utilize only gallery set; (2) they  optimize all gallery samples in the initial ranking list while we focus on the first $L$ samples instead ($L=200$). 
Although there are a collection of probe samples and a collection of gallery samples in currently person re-id datasets, the input of person re-id system is often only a probe sample and a collection of gallery samples in practical applications, which is fully taken into account in the proposed method.
What’s more, in contrast to the proposed method, these two methods introduce more information to optimize all gallery samples and thus are more likely to achieve better performance. However, from the above comparison results we can see that these two methods do not have significant performance advantages on Rank-1 and mAP compared to the proposed method on all four datasets. Meanwhile, compared to the proposed method, these methods using both probe and gallery samples to optimize all gallery samples mean higher computation complexity and space complexity. It can be seen from Tables~\ref{tab6}, \ref{tab7} and \ref{tab8} that the proposed method has an obvious advantage on the run time. Particularly, we conduct experiments for the proposed method and ECN method on a PC ($3.4$GHz Intel Core CPU, $32$GB RAM) and a server ($2.6$GHz Intel Xeon CPU, $128$GB RAM), respectively. As shown in Tables~\ref{tab10}, it takes a very long time to run ECN on the PC, since almost $100\%$ memory is occupied during the execution of the ECN method on the Market1501 and DukeMTMC datasets, and ECN can be run in a reasonable span of time only on the server with more memory; in contrast, the proposed method can be run efficiently both on the PC or server, showing a low space complexity of the proposed method.
Above all, the proposed method is more efficient and more practical.

\section{Discussion}
\subsection{Performance on mAP}
\begin{figure}
\begin{center}
   \includegraphics[width=0.96\linewidth, height=0.76\linewidth]{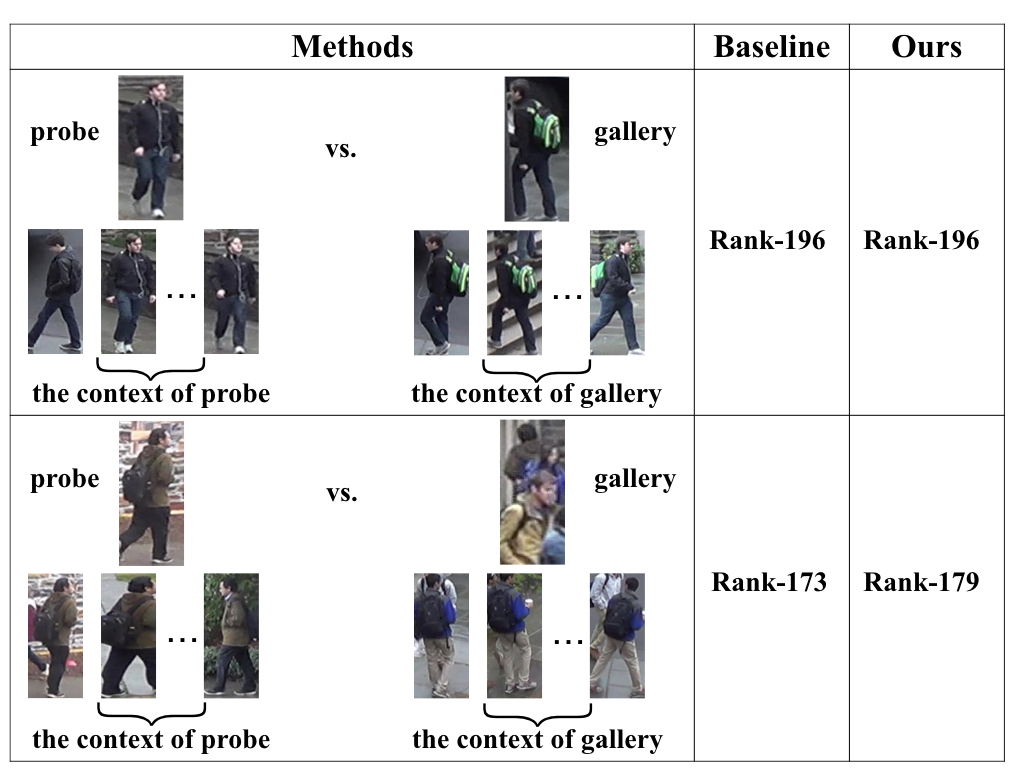}
\end{center}
   \caption{Two examples of re-identification results not being improved by the proposed method. The PCB method \cite{sun2018beyond} is used as the baseline.}
\label{fig6}
\end{figure}

It can be seen from the experiments that the proposed method can effectively improve the rank-1 matching rate by $1.1\%-11.0\%$ and the mAP scores by $5.7\%-15.4\%$ of the baseline methods on the Market1501, DukeMTMC, CUHK03 and MARS datasets, and has advantages on evaluation metrics and run time compared to the other person re-id methods. However, there remains one open issue.  
Compared to the other post-processing person re-id methods, the mAP of the baseline method is not improved obviously by the proposed method. We only optimize the ranking positions of top-$200$ samples in consideration of the efficiency in the proposed method, relative to other post-processing methods where all samples are optimized, resulting in the relatively low mAP of the proposed method. We can see in Fig.~\ref{fig7} that if more gallery samples are optimized by increasing $L$, better results on mAP can be achieved by the proposed method. 
Beyond that, there is a more fundamental reason. For a specific probe, there is likely to be a large variation in appearance with the probe sample for the positive gallery sample with lower ranking in the initial ranking list. As a result, the samples in context of the probe sample (gallery sample) are not similar to the gallery sample (probe sample) and the proposed method might fail to improve the gallery sample's initial ranking position with the help of this contextual information. The results of this positive gallery sample with lower ranking in the initial ranking list are not improved by the proposed method, which leads to the relatively low mAP. Two examples of the failure cases are shown in Fig.~\ref{fig6}. For such cases, how to improve the ranking results and thereby improve the mAP is a problem that needs to be solved in future. One can for instance try to use more extra information such as the partner's information to improve the ranking results. 

\subsection{Evaluation on small-scale dataset}
\begin{figure}
\begin{center}
   \includegraphics[width=0.9\linewidth, height=0.6\linewidth]{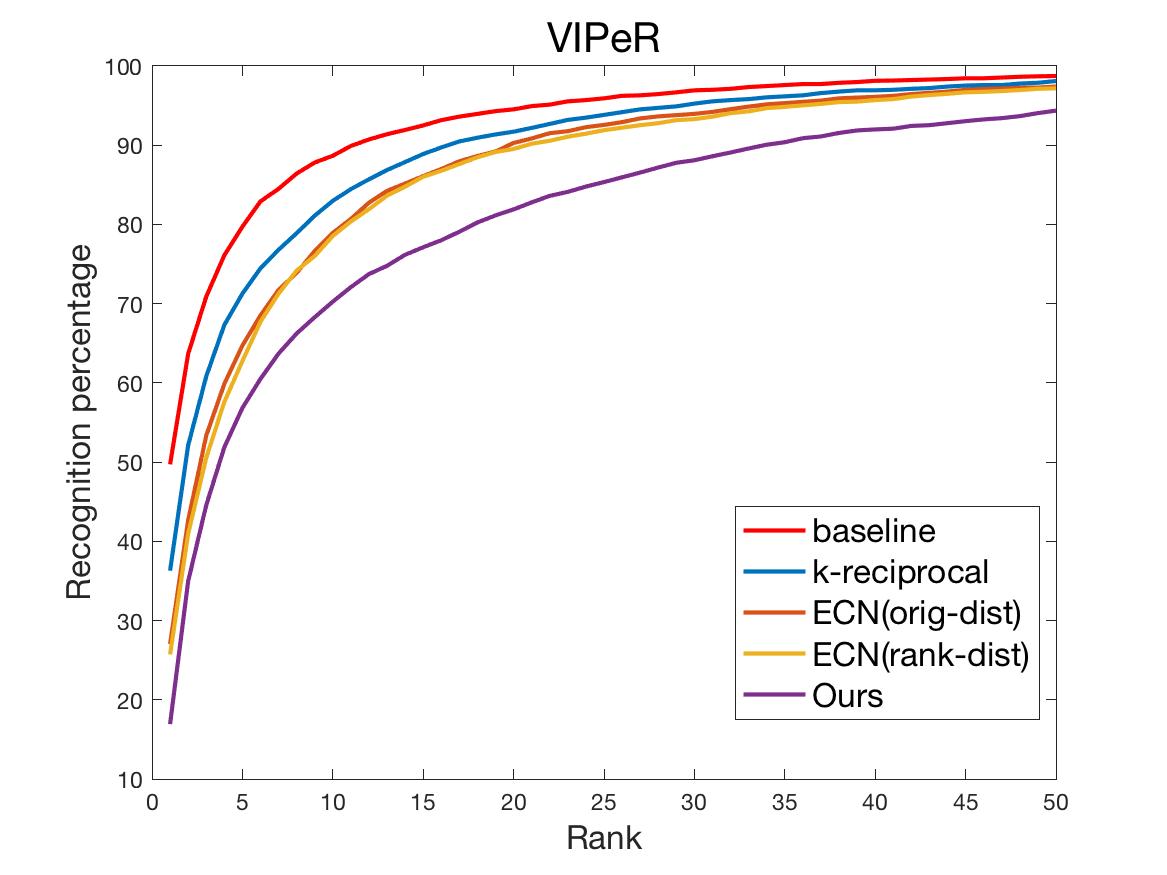}
\end{center}
   \caption{Comparison with the baseline method LOMO+XQDA \cite{liao2015person} on the VIPeR dataset.}
\label{fig2}
\end{figure}

The proposed method effectively improves the performance of person re-id on four large-scale datasets, which benefits from abundant contextual information of the sample provided by these datasets. And how has the proposed method performed on small-scale dataset in which there is only one person image for each of person IDs from one camera? Fig.~\ref{fig2} shows the results on the small-scale dataset VIPeR \cite{Douglas2008Viewpoint}. It can bee seen that the performance of baseline is not improved by the proposed method, similarly, the performance is also not improved by k-reciprocal \cite{zhong2017re} and ECN \cite{saquib2018pose} methods. In these methods, the performance enhancement is achieved by utilizing the contextual information. 
However, there is only one image for each person ID from one camera and the individual differences on appearance are considerable on small-scale dataset such as VIPeR, with the result that the effective contextual information is very limited and the performance enhancement can not be effectively achieved by the proposed method, k-reciprocal and ECN methods  on these datasets. In addition, we can see from Fig.~\ref{fig2} that the proposed method experiences the largest performance decline compared to the k-reciprocal and ECN methods. It indicates that the proposed method makes the best use of the contextual information. As such, the performance is not very good on VIPeR dataset with the limited contextual information. 
In future, we can consider using other information to improve the performance on small-scale dataset.

\section{Conclusion}
In this paper, we propose a generic post-processing unsupervised person re-id method by which most of the content-based person re-id methods can improve their performance even further. The proposed method is based on the assumption that the target sample pair are similar to each other if any one of target sample's context is similar to the remaining one. In doing so, the pairwise similarity is computed based on the consideration of the dataset manifold structure. The proposed method is effective and efficient. It can well handle the real-world large scale person re-id task in both accuracy performance and computation complexity. Experimental results on four large-scale person re-id datasets demonstrate the superiority of the proposed method compared to the state-of-the-art person re-id methods. It is worth mentioning that the proposed method has significant advantage in computing efficiency compared to other post-processing person re-id methods.

For future development, we will investigate how to refine the match results of positive sample pairs having a large variations to each other in appearance, and how to improve the performance of person re-id on small-scale dataset.


%

\section*{Acknowledgment}

This work was supported by the National Key R\&D Program of China under Grant 2017YFC0803505 and the National Natural Science Foundation of China under Grant NSFC 61906194. 
This work was partly supported by the National Key R\&D Program of China under Grant 25904.

\ifCLASSOPTIONcaptionsoff
  \newpage
\fi



%

\bibliography{mybibfile}

%





\end{document}